\documentclass{article}

\PassOptionsToPackage{square,numbers,comma,sort&compress}{natbib}

\usepackage[final]{neurips_2024_mlsb}

\usepackage[utf8]{inputenc} 
\usepackage[T1]{fontenc}    
\usepackage[hidelinks]{hyperref}       
\usepackage{url}            
\usepackage{booktabs}       
\usepackage{amsmath,amsfonts}       
\usepackage{nicefrac}       
\usepackage{microtype}      
\usepackage{xcolor}         
\usepackage[pdftex]{graphicx}
\usepackage{tabularx}
\usepackage{float}
\usepackage{easytable}

\hypersetup{
    colorlinks,
    linkcolor={red!50!black},
    citecolor={blue!50!black},
    urlcolor={blue!80!black}
}
\newcolumntype{Y}{>{\centering\arraybackslash}X}

\title{Similarity-Quantized Relative Difference Learning for Improved Molecular Activity Prediction}

\author{%
  \textbf{Karina Zadorozhny} \quad \textbf{Kangway V.~Chuang} \quad \textbf{Bharath Sathappan} \\
  \textbf{Ewan Wallace} \quad \textbf{Vishnu Sresht} \quad \textbf{Colin A.~Grambow} \\
  Prescient Design, Genentech \\
  South San Francisco, CA 94080 \\
  \texttt{\{zadorozhny.karina,grambow.colin\}@gene.com} \\
}

\begin{document}

\maketitle

\begin{abstract}

Accurate prediction of molecular activities is crucial for efficient drug discovery, yet remains challenging due to limited and noisy datasets. We introduce Similarity-Quantized Relative Learning (SQRL), a learning framework that reformulates molecular activity prediction as relative difference learning between structurally similar pairs of compounds. SQRL uses precomputed molecular similarities to enhance training of graph neural networks and other architectures, and significantly improves accuracy and generalization in low-data regimes common in drug discovery. We demonstrate its broad applicability and real-world potential through benchmarking on public datasets as well as proprietary industry data. Our findings demonstrate that leveraging similarity-aware relative differences provides an effective paradigm for molecular activity prediction.

\end{abstract}

\section{Introduction and Background}
\label{sec:intro}

The ability to predict molecular activity is critical for small molecule drug discovery. However, experimental data for training machine learning models are often limited and noisy, making robust generalization challenging. While deep learning approaches have enabled rich representations from chemical structures~\citep{Atz2021geometricdeeplearning, gilmer2017neuralmessagepassingquantum,yang2019analyzinglearnedmolecularrepresentations,Xiong2020AttentiveFP, liu2022sphericalmessagepassing3d,schütt2017schnetcontinuousfilterconvolutionalneural,satorras2022enequivariantgraphneural,Wang2019smilesbert, honda2019smilestransformerpretrainedmolecular, liu2019robertarobustlyoptimizedbert, chithrananda2020chembertalargescaleselfsupervisedpretraining, ross2022largescalechemicallanguagerepresentations,Chuang2020}, most focus on predicting absolute property values, ignoring valuable information in relationships between structurally similar molecules.

Drawing inspiration from medicinal chemists~\citep{Thomas2004-yg}, who examine how specific structural modifications influence properties relative to a parent compound or a matched molecular pair~\citep{Yang2023-matchedmolecularpairs}, we introduce \textbf{S}imilarity-\textbf{Q}uantized \textbf{R}elative \textbf{L}earning (\textbf{SQRL})---a training and evaluation framework that reformulates property prediction as learning relative differences between nearby compounds. 

The key contributions of our work are:

\begin{itemize}
     \item We introduce a robust \textbf{similarity-thresholded learning approach} that significantly enhances model performance by focusing on \textbf{predicting property differences between the most informative compound pairs}. This allows effective learning from limited and noisy data.
    \item Our analysis across molecular distance metrics and thresholds demonstrates that \textbf{similarity-aware dataset matching outperforms indiscriminate pairing of all inputs}, as it more effectively leverages local structural information.
    \item Extensive \textbf{benchmarking} demonstrating the benefits of SQRL \textbf{across diverse state-of-the-art network architectures} and multiple activity prediction datasets, including publicly available activity cliff prediction tasks and real-world industry datasets.
\end{itemize}

\section{Related Work}
\label{sec:relwork}

\paragraph{Molecular property prediction.} Significant research has focused on methods for molecular property and activity prediction, including recent work on graph neural networks (GNNs)~\citep{Atz2021geometricdeeplearning, gilmer2017neuralmessagepassingquantum,yang2019analyzinglearnedmolecularrepresentations,Xiong2020AttentiveFP, liu2022sphericalmessagepassing3d,schütt2017schnetcontinuousfilterconvolutionalneural,satorras2022enequivariantgraphneural} along with chemical language models~\citep{Wang2019smilesbert, honda2019smilestransformerpretrainedmolecular, liu2019robertarobustlyoptimizedbert, chithrananda2020chembertalargescaleselfsupervisedpretraining, ross2022largescalechemicallanguagerepresentations} that learn meaningful molecular representations from molecular data~\citep{Chuang2020}. Despite these advances, simpler tree-based models often outperform more complex neural approaches due to limited data availability and inherent modeling challenges~\citep{Deng2023-ve, Xia2023understandinglimitationsofdeep}. 

\paragraph{Activity cliff prediction.} Activity cliffs refer to pairs of molecules with high structural similarity but significantly different activity levels. Previous approaches have addressed this challenging problem by representing molecular graphs as images~\citep{chen2024maskmol,iqbal2021prediction}, applying graph convolutional networks (GCNs) to matched molecular pairs~\citep{park_acgcn}, or using chemical reaction information~\citep{horvath2016prediction}. Many of these methods have formulated the problem as a classification task, aiming to identify whether a given molecular pair exhibits an activity cliff~\citep{Dablander2023} or as a standard regression task that relies on learning the discontinuous chemical space directly from the data~\citep{Tilborg2022moleculeace}. Our approach focuses on predicting the difference in potency values between any pair of similar molecules, providing a more versatile solution that can also be applied to standard potency prediction.

\paragraph{Metric, similarity, and few-shot learning.} Pairwise learning approaches have been widely adopted in fields like ranking, metric, and similarity learning~\citep{Kulis2013, Liu2011}. Notably, few-shot learning approaches have been developed for molecular property prediction for improving generalization in low-data regimes~\citep{Altae-Tran2017-OneShot, Vella2023-FS, Stanley-FsMol}. Additionally, pairwise data matching has proven effective in implicit guidance of generative models for drug design. \citep{tagasovska2024implicitlyguideddesignpropen}.

\paragraph{Relative prediction.} Recently, pairwise learning has been applied to regression tasks. \citeauthor{Wetzel2022-A} used Siamese networks to predict differences between all data points in both supervised and unsupervised settings as a way of producing ensembles of predictions and uncertainty estimates~\citep{Wetzel2022-A, Wetzel2022-B}. This approach has been extended to classification tasks for tree-based models~\citep{Belaid2024}. \citeauthor{Tynes2021} applied the concept of pairwise learning to computational chemistry, analyzing the performance of random forest models trained on all pairs of inputs points~\citep{Tynes2021}. Similarly, \citeauthor{Fralish2023} trained a \mbox{D-MPNN} model on paired compounds and observed improvements in ADME property predictions~\citep{Fralish2023} and molecule selection in an active learning setting~\citep{Fralish2024activelearningpairs}.

Unlike previous approaches that primarily focus on absolute property predictions or indiscriminate pairwise learning, our work introduces a similarity-thresholded framework that emphasizes learning from the most informative compound pairs.

\section{Similarity-Thresholded Relative Representation}
\label{sec:methods}

\paragraph{Problem formulation.}
We formulate the relative prediction task as follows. Given a dataset of molecular structures $\mathcal{D} = \{(x_i, y_i)\}_{i=1}^N$, where $x_i$ represents molecule $i$ and $y_i \in \mathbb{R}$ denotes its corresponding property value, our goal is to learn a function $f: \mathcal{X} \times \mathcal{X} \rightarrow \mathbb{R}$ that predicts the relative difference in property values between two molecules. Formally, for any pair of molecules $(x_i, x_j)$, we aim to predict the relative difference $\Delta y_{ij} = y_i - y_j$.

\paragraph{Dataset matching.}
To train models on the relative prediction task, we construct a new dataset $\mathcal{D}_{\text{rel}}$ by considering pairs of molecules in the original dataset $\mathcal{D}$. To focus on the most informative comparisons, we restrict the pairs to those within a certain threshold of structural similarity, as measured by a predefined similarity metric (e.g., Tanimoto similarity). This allows models to learn from local differences in chemical space where relative changes in property values are most meaningful. We define $\mathcal{D}_{\text{rel}}$ as:
\begin{equation}
\mathcal{D}_{\text{rel}} = \left\{ \left( (x_i, x_j), \Delta y_{ij} \right) \mid x_i, x_j \in \mathcal{D}, d(x_i, x_j) \leq \alpha \right\}
\end{equation}
where $d: \mathcal{X} \times \mathcal{X} \rightarrow \mathbb{R}_{\geq 0}$ is a distance function in the input space $\mathcal{X}$ and $\alpha \in \mathbb{R}_{> 0}$ is a distance threshold. See Appendix~\ref{appendix:pairs} for examples of paired structures.

The choice of $\alpha$ is critical, as it involves a trade-off between the quantity and relevance of generated pairs. We propose selecting $\alpha$ based on the distribution of distances in the training data, specifically by choosing a threshold smaller than the average pairwise distance. This approach can help form more informative pairs by focusing on molecules with greater structural similarity and relevance.

\paragraph{Relative representation.}
We define \( g: \mathcal{X} \rightarrow \mathbb{R}^d \) as a mapping function that converts a molecular compound from the input space \(\mathcal{X}\) into a \(d\)-dimensional real-valued vector. This can be either a learnable model (such as a graph neural network), a pre-trained model, or a fixed molecular fingerprinting algorithm. Next, a machine learning model \( f: \mathbb{R}^d \rightarrow \mathbb{R} \) uses the difference between molecular representations generated by \( g \) to predict the relative differences in properties.

We optimize parameters $\theta$ of $f$ and $g$ (if learnable) by minimizing:
\begin{equation}
\min_\theta \mathcal{L}(\theta) = \min_\theta \sum_{((x_i, x_j), \Delta y_{ij}) \in \mathcal{D}{\text{rel}}} \ell \big( f\left( g(x_i) - g(x_j) \right), \Delta y_{ij} \big)
\end{equation}
where $\ell$ is mean squared error loss. For a new molecule $x_\text{new}$, we compute the prediction $\hat{y}_\text{new}$ as:
\begin{equation}
\hat{y}_\text{new} = \frac{1}{n} \sum_{x_i \in \text{NN}_n (x_\text{new})} y_i + f\left( g(x_i) - g(x_\text{new}) \right)
\end{equation}
where $\text{NN}_n (x_\text{new})$ denotes the set of $n$ molecules from the training data $\mathcal{D}$ that are nearest to $x_\text{new}$ as determined by the distance function $d(x_i, x_\text{new})$. Unless otherwise specified, we set $n = 1$.

\section{Experimental Results}
\label{sec:experiments}

\begin{figure}[tb]%
    \centering%
    \includegraphics[width=\linewidth]{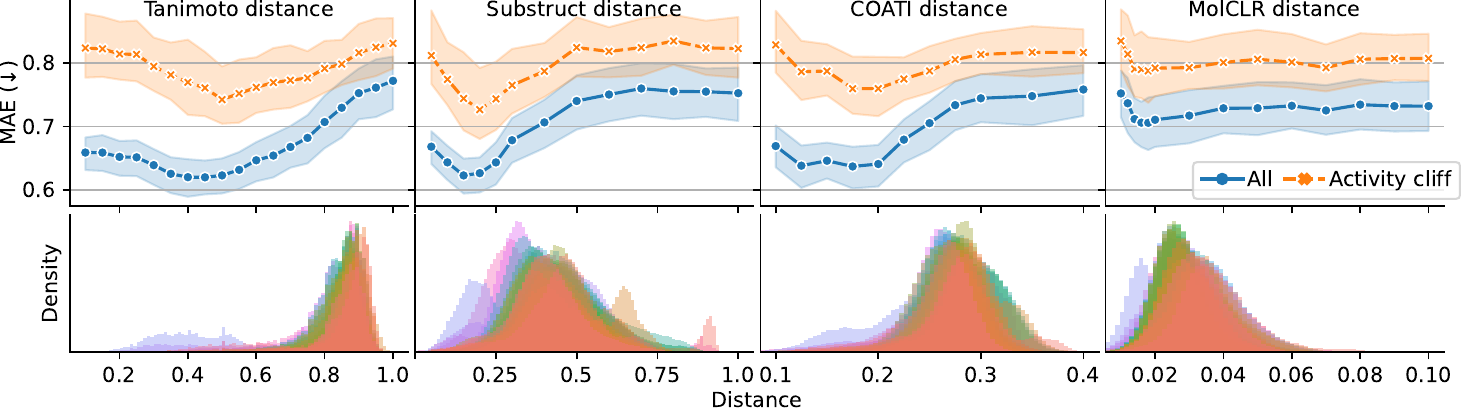}%
    \caption{\textbf{Leveraging local structural information enhances predictive performance.} \emph{Top}: Incorporating neighbors only up to a certain distance threshold $\alpha$ improves MAE ($\downarrow$). \emph{Bottom}: Pairwise distance distributions of training data (overlaid for all 30 MoleculeACE tasks) with greater skewness and kurtosis yield the best performance and a wider range of acceptable values of $\alpha$.}%
    \label{fig:dist_sweep}%
\end{figure}%

\begin{table}[tb]
\centering
\caption{\textbf{Spearman's $\rho$ ($\uparrow$) comparison of predictive performance.} Spearman's~$\rho$ with standard deviation across tasks for Standard and SQRL methods using Tanimoto distance with $\alpha = 0.7$. Uni-Mol was evaluated on a subset of tasks due to computational constraints. See Appendix~\ref{appendix:models} for more details and MAE scores (Table~\ref{tab:mae_results}).}
\label{tab:spearman_results}
\setlength{\tabcolsep}{2pt}
\begin{tabularx}{\linewidth}{@{}lYYYYYY@{}}
\toprule
& \multicolumn{2}{c}{\textbf{MoleculeACE}} & \multicolumn{2}{c}{\textbf{MoleculeACE-Cliff}} & \multicolumn{2}{c}{\textbf{Internal Targets}} \\
\cmidrule(lr){2-3} \cmidrule(lr){4-5} \cmidrule(lr){6-7}
\textbf{Model} & Standard & SQRL & Standard & SQRL & Standard & SQRL \\
\midrule
\emph{Baselines} \\
\cmidrule(r){1-1}
XGBoost & \textbf{0.79 $\pm$ 0.10} & 0.76 $\pm$ 0.08 & \textbf{0.72 $\pm$ 0.18} & 0.64 $\pm$ 0.15 & \textbf{0.73 $\pm$ 0.14} & 0.65 $\pm$ 0.16 \\
RF & \textbf{0.80 $\pm$ 0.09} & 0.77 $\pm$ 0.07 & \textbf{0.72 $\pm$ 0.18} & 0.68 $\pm$ 0.14 & \textbf{0.70 $\pm$ 0.13} & 0.68 $\pm$ 0.14 \\
KNN & 0.66 $\pm$ 0.10 & \textbf{0.67 $\pm$ 0.13} & 0.57 $\pm$ 0.19 & \textbf{0.59 $\pm$ 0.23} & \textbf{0.52 $\pm$ 0.14} & \textbf{0.52 $\pm$ 0.14} \\
MLP & 0.32 $\pm$ 0.17 & \textbf{0.73 $\pm$ 0.09} & 0.23 $\pm$ 0.16 & \textbf{0.62 $\pm$ 0.16} & 0.39 $\pm$ 0.14 & \textbf{0.59 $\pm$ 0.20} \\
\midrule
\emph{GNNs} \\
\cmidrule(r){1-1}
AttentiveFP & 0.52 $\pm$ 0.20 & \textbf{0.77 $\pm$ 0.09} & 0.43 $\pm$ 0.19 & \textbf{0.67 $\pm$ 0.17} & 0.49 $\pm$ 0.13 & \textbf{0.66 $\pm$ 0.19} \\
GINE & 0.33 $\pm$ 0.19 & \textbf{0.76 $\pm$ 0.09} & 0.29 $\pm$ 0.21 & \textbf{0.68 $\pm$ 0.18} & 0.40 $\pm$ 0.22 & \textbf{0.61 $\pm$ 0.17} \\
PNA & 0.51 $\pm$ 0.18 & \textbf{0.72 $\pm$ 0.08} & 0.41 $\pm$ 0.17 & \textbf{0.61 $\pm$ 0.18} & 0.52 $\pm$ 0.12 & \textbf{0.64 $\pm$ 0.15} \\
MolCLR & 0.35 $\pm$ 0.22 & \textbf{0.77 $\pm$ 0.09} & 0.28 $\pm$ 0.20 & \textbf{0.66 $\pm$ 0.18} & 0.39 $\pm$ 0.27 & \textbf{0.68 $\pm$ 0.17} \\
\midrule
\emph{Transformers} \\
\cmidrule(r){1-1}
COATI & 0.69 $\pm$ 0.11 & \textbf{0.74 $\pm$ 0.09} & 0.59 $\pm$ 0.15 & \textbf{0.64 $\pm$ 0.16} & 0.52 $\pm$ 0.26 & \textbf{0.61 $\pm$ 0.18} \\
Uni-Mol & 0.26 $\pm$ 0.19 & \textbf{0.69 $\pm$ 0.10} & 0.19 $\pm$ 0.20 & \textbf{0.57 $\pm$ 0.18} & 0.40 $\pm$ 0.16 & \textbf{0.51 $\pm$ 0.14} \\
SAFE-GPT & 0.61 $\pm$ 0.20 & \textbf{0.71 $\pm$ 0.08} & 0.56 $\pm$ 0.22 & \textbf{0.61 $\pm$ 0.14} & 0.58 $\pm$ 0.14 & \textbf{0.59 $\pm$ 0.15} \\
\bottomrule
\end{tabularx}
\end{table}

\subsection{Experimental setup}
\label{sec:experimental_setup}

To understand the generality of SQRL, we conducted extensive evaluations across a diverse set of models and molecular activity datasets. Each model was trained to predict absolute property values directly (Standard) or to predict relative differences between selected molecule pairs (SQRL). More details about models, hyperparameter selection, and datasets can be found in Appendices~\ref{appendix:models}--\ref{appendix:pairs}.

\paragraph{Models.} We benchmarked baselines (RF~\citep{Breiman2001-nj}, XGBoost~\citep{chen2016xgboost}, KNN) using Morgan fingerprints~\citep{Rogers2010-vf} with RDKit features~\citep{Landrum2016RDKit2016_09_4}, MLP with Morgan fingerprints, GNNs (AttentiveFP~\citep{Xiong2020AttentiveFP}, GINE~\citep{hu2020gine}, PNA~\citep{corso2020principalneighbourhoodaggregationgraph}, MolCLR~\citep{wang2022molclr}), and transformer models (COATI~\citep{kaufman2023coati}, SAFE-GPT~\citep{noutahi2023gottasafenewframework}, and Uni-Mol~\citep{zhou2023unimol}).

\paragraph{Distance metrics.} Several distance metrics were used to create training data pairs and obtain the closest training molecules for inference. We evaluated Tanimoto distances between Morgan fingerprints~\citep{Rogers2010-vf}, Tanimoto distances between substructure count vectors~\citep{ehrlich2012}, as well as Euclidean distances using COATI~\citep{kaufman2023coati}, Uni-Mol, and MolCLR embeddings~\citep{wang2022molclr} (see~\ref{appendix:dist_sweep}).

\paragraph{Datasets.} We evaluated our approach on 30 activity prediction tasks (pEC\textsubscript{50}/p$K_i$) for ChEMBL targets using the MoleculeACE dataset~\citep{Tilborg2022moleculeace}. This dataset includes challenging activity cliff molecules (MoleculeACE-Cliff)---structurally similar compounds with large activity differences---providing a crucial test of local generalizability (see~\ref{appendix:pairs}). Additionally, we evaluated models on 5 proprietary drug discovery projects (Internal Targets) to assess the real-world applicability of our approach. Models are trained on single tasks and results are reported aggregated across tasks.

\subsection{Leveraging local structural information improves learning}
\label{sec:sim_thresh}

A key hypothesis of SQRL is that not all possible relative pairs are equally informative, and training on all pairwise comparisons as done previously~\citep{Fralish2023,Tynes2021,Belaid2024} may overemphasize global relationships at the expense of local consistency. We analyzed pairwise distance distributions using various distance metrics and found that some distributions exhibit significant left-skew and/or high kurtosis (Figure~\ref{fig:dist_sweep} and Appendix Figure~\ref{fig:dist_sweep_appendix}). We hypothesize that these highly similar compounds contain the most informative signal. To test this assumption, we trained MLP models on top of Morgan fingerprint features across a range of similarity thresholds for each of the metrics (Figure~\ref{fig:dist_sweep}). We observed that incorporating neighbors up to a certain threshold provides a pronounced improvement in performance, but beyond this point, including more dissimilar pairs degrades performance. Moreover, performance is best for distance metrics with the desired distribution characteristics (e.g., Tanimoto and COATI). These results demonstrate the benefit of our similarity-thresholded approach: a smaller distance threshold $\alpha$ yields fewer but potentially more informative pairs, while a larger threshold increases pair count but may introduce less relevant comparisons. Our findings support the hypothesis that focusing on the most similar pairs effectively leverages local structural information (Appendix Figure~\ref{fig:dist_to_train}), leading to improved predictive performance across various model architectures and datasets.

\subsection{SQRL consistently improves predictive performance across all neural network architectures}
\label{sec:performance_improvement}

To understand which model types benefit from this approach, we performed extensive benchmarking of molecular property prediction models trained using both the standard absolute prediction objective and SQRL. While the baseline models did not significantly benefit from relative training, all deep learning architectures exhibit consistent improvement of the Spearman rank correlation across all datasets when trained with SQRL (Table~\ref{tab:spearman_results}). We observe significant improvements for all GNN architectures and especially for the pre-trained transformer-based models---0.57 and 0.43 point improvement for COATI and Uni-Mol, respectively. Notably, substantial improvements are observed on the MoleculeACE-Cliff subset, highlighting SQRL's ability to capture fine-grained structural differences that significantly impact molecular properties. Evaluation on internal targets shows that the observed benefits are transferable to real-world scenarios and underscore the robustness and broad applicability of this approach.

The lack of improvement for XGBoost, RF, and KNN with SQRL is notable. This may indicate that the simple difference fingerprint representation is not sufficient for these models as it forces these models to only learn from substructures that differ between a pair of molecules without taking into account the rest of the molecular structure of both molecules. A higher fidelity representation, such as concatenating the full molecular fingerprints to the difference representation, may overcome this limitation. Even so, most modern deep learning methods do not outperform conventional baselines (with or without SQRL). The small size of the training data for each task likely contributes to this result. However, we expect that neural network approaches may outperform baselines with additional tuning (e.g., more rigorous hyperparameter selection, choosing a more optimal distance threshold, or redesigning the training objective).

\section{Conclusions and Future Directions}

Overall, our work demonstrates consistent improvements for neural networks in molecular activity prediction using SQRL, particularly in capturing molecular activity cliffs. Our method shows promise in learning from pairwise differences, potentially offering a more nuanced understanding of structure-activity relationships. The limitations of the current work include the assumption that meaningful distance metrics are available. Future research could focus on refining SQRL for applications where similarity measures are less well-defined or more challenging to establish.

{
\small
\bibliography{main.bib}

\begin{thebibliography}{47}
\providecommand{\natexlab}[1]{#1}
\providecommand{\url}[1]{\texttt{#1}}
\expandafter\ifx\csname urlstyle\endcsname\relax
  \providecommand{\doi}[1]{doi: #1}\else
  \providecommand{\doi}{doi: \begingroup \urlstyle{rm}\Url}\fi

\bibitem[Atz et~al.(2021)Atz, Grisoni, and Schneider]{Atz2021geometricdeeplearning}
Kenneth Atz, Francesca Grisoni, and Gisbert Schneider.
\newblock Geometric deep learning on molecular representations.
\newblock \emph{Nature Machine Intelligence}, 3\penalty0 (12):\penalty0 1023--1032, 2021.
\newblock URL \url{https://doi.org/10.1038/s42256-021-00418-8}.

\bibitem[Gilmer et~al.(2017)Gilmer, Schoenholz, Riley, Vinyals, and Dahl]{gilmer2017neuralmessagepassingquantum}
Justin Gilmer, Samuel~S. Schoenholz, Patrick~F. Riley, Oriol Vinyals, and George~E. Dahl.
\newblock Neural message passing for quantum chemistry.
\newblock In \emph{Proceedings of the 34th International Conference on Machine Learning - Volume 70}, ICML'17, pages 1263--1272, 2017.
\newblock URL \url{https://proceedings.mlr.press/v70/gilmer17a/gilmer17a.pdf}.

\bibitem[Yang et~al.(2019)Yang, Swanson, Jin, Coley, Eiden, Gao, Guzman-Perez, Hopper, Kelley, Mathea, Palmer, Settels, Jaakkola, Jensen, and Barzilay]{yang2019analyzinglearnedmolecularrepresentations}
Kevin Yang, Kyle Swanson, Wengong Jin, Connor Coley, Philipp Eiden, Hua Gao, Angel Guzman-Perez, Timothy Hopper, Brian Kelley, Miriam Mathea, Andrew Palmer, Volker Settels, Tommi Jaakkola, Klavs Jensen, and Regina Barzilay.
\newblock Analyzing learned molecular representations for property prediction.
\newblock \emph{Journal of Chemical Information and Modeling}, 59\penalty0 (8):\penalty0 3370--3388, 2019.
\newblock URL \url{https://doi.org/10.1021/acs.jcim.9b00237}.

\bibitem[Xiong et~al.(2020)Xiong, Wang, Liu, Zhong, Wan, Li, Li, Luo, Chen, Jiang, and Zheng]{Xiong2020AttentiveFP}
Zhaoping Xiong, Dingyan Wang, Xiaohong Liu, Feisheng Zhong, Xiaozhe Wan, Xutong Li, Zhaojun Li, Xiaomin Luo, Kaixian Chen, Hualiang Jiang, and Mingyue Zheng.
\newblock Pushing the boundaries of molecular representation for drug discovery with the graph attention mechanism.
\newblock \emph{Journal of Medicinal Chemistry}, 63\penalty0 (16):\penalty0 8749--8760, 2020.
\newblock URL \url{https://doi.org/10.1021/acs.jmedchem.9b00959}.

\bibitem[Liu et~al.(2022)Liu, Wang, Liu, Lin, Zhang, Oztekin, and Ji]{liu2022sphericalmessagepassing3d}
Yi~Liu, Limei Wang, Meng Liu, Yuchao Lin, Xuan Zhang, Bora Oztekin, and Shuiwang Ji.
\newblock Spherical message passing for {3D} molecular graphs.
\newblock In \emph{International Conference on Learning Representations}, 2022.
\newblock URL \url{https://openreview.net/forum?id=givsRXsOt9r}.

\bibitem[Sch\"{u}tt et~al.(2017)Sch\"{u}tt, Kindermans, Sauceda, Chmiela, Tkatchenko, and M\"{u}ller]{schütt2017schnetcontinuousfilterconvolutionalneural}
K.~T. Sch\"{u}tt, P.-J. Kindermans, H.~E. Sauceda, S.~Chmiela, A.~Tkatchenko, and K.-R. M\"{u}ller.
\newblock {SchNet}: A continuous-filter convolutional neural network for modeling quantum interactions.
\newblock In \emph{Proceedings of the 31st International Conference on Neural Information Processing Systems}, NIPS'17, pages 992--1002, Red Hook, NY, USA, 2017. Curran Associates Inc.
\newblock URL \url{https://proceedings.neurips.cc/paper_files/paper/2017/file/303ed4c69846ab36c2904d3ba8573050-Paper.pdf}.

\bibitem[Satorras et~al.(2022)Satorras, Hoogeboom, and Welling]{satorras2022enequivariantgraphneural}
Victor~Garcia Satorras, Emiel Hoogeboom, and Max Welling.
\newblock E(n) equivariant graph neural networks, 2022.
\newblock URL \url{https://arxiv.org/abs/2102.09844}.

\bibitem[Wang et~al.(2019)Wang, Guo, Wang, Sun, and Huang]{Wang2019smilesbert}
Sheng Wang, Yuzhi Guo, Yuhong Wang, Hongmao Sun, and Junzhou Huang.
\newblock {SMILES-BERT}: Large scale unsupervised pre-training for molecular property prediction.
\newblock In \emph{Proceedings of the 10th ACM International Conference on Bioinformatics, Computational Biology and Health Informatics}, BCB '19, pages 429--436, New York, NY, USA, 2019. Association for Computing Machinery.
\newblock URL \url{https://doi.org/10.1145/3307339.3342186}.

\bibitem[Honda et~al.(2019)Honda, Shi, and Ueda]{honda2019smilestransformerpretrainedmolecular}
Shion Honda, Shoi Shi, and Hiroki~R. Ueda.
\newblock {SMILES Transformer}: Pre-trained molecular fingerprint for low data drug discovery, 2019.
\newblock URL \url{https://arxiv.org/abs/1911.04738}.

\bibitem[Liu et~al.(2019)Liu, Ott, Goyal, Du, Joshi, Chen, Levy, Lewis, Zettlemoyer, and Stoyanov]{liu2019robertarobustlyoptimizedbert}
Yinhan Liu, Myle Ott, Naman Goyal, Jingfei Du, Mandar Joshi, Danqi Chen, Omer Levy, Mike Lewis, Luke Zettlemoyer, and Veselin Stoyanov.
\newblock {RoBERTa}: A robustly optimized {BERT} pretraining approach, 2019.
\newblock URL \url{https://arxiv.org/abs/1907.11692}.

\bibitem[Chithrananda et~al.(2020)Chithrananda, Grand, and Ramsundar]{chithrananda2020chembertalargescaleselfsupervisedpretraining}
Seyone Chithrananda, Gabriel Grand, and Bharath Ramsundar.
\newblock {ChemBERTa}: Large-scale self-supervised pretraining for molecular property prediction, 2020.
\newblock URL \url{https://arxiv.org/abs/2010.09885}.

\bibitem[Ross et~al.(2022)Ross, Belgodere, Chenthamarakshan, Padhi, Mroueh, and Das]{ross2022largescalechemicallanguagerepresentations}
Jerret Ross, Brian Belgodere, Vijil Chenthamarakshan, Inkit Padhi, Youssef Mroueh, and Payel Das.
\newblock Large-scale chemical language representations capture molecular structure and properties.
\newblock \emph{Nature Machine Intelligence}, 4\penalty0 (12):\penalty0 1256--1264, 2022.
\newblock URL \url{https://doi.org/10.1038/s42256-022-00580-7}.

\bibitem[Chuang et~al.(2020)Chuang, Gunsalus, and Keiser]{Chuang2020}
Kangway~V. Chuang, Laura~M. Gunsalus, and Michael~J. Keiser.
\newblock Learning molecular representations for medicinal chemistry.
\newblock \emph{Journal of Medicinal Chemistry}, 63\penalty0 (16):\penalty0 8705--8722, 2020.
\newblock URL \url{https://doi.org/10.1021/acs.jmedchem.0c00385}.

\bibitem[Thomas(2004)]{Thomas2004-yg}
Gareth Thomas.
\newblock \emph{Fundamentals of Medicinal Chemistry}.
\newblock John Wiley \& Sons, Chichester, England, 1st edition, March 2004.

\bibitem[Yang et~al.(2023)Yang, Shi, Fu, Lu, Hou, and Cao]{Yang2023-matchedmolecularpairs}
Ziyi Yang, Shaohua Shi, Li~Fu, Aiping Lu, Tingjun Hou, and Dongsheng Cao.
\newblock Matched molecular pair analysis in drug discovery: Methods and recent applications.
\newblock \emph{Journal of Medicinal Chemistry}, 66\penalty0 (7):\penalty0 4361--4377, 2023.
\newblock URL \url{https://doi.org/10.1021/acs.jmedchem.2c01787}.

\bibitem[Deng et~al.(2023)Deng, Yang, Wang, Ojima, Samaras, and Wang]{Deng2023-ve}
Jianyuan Deng, Zhibo Yang, Hehe Wang, Iwao Ojima, Dimitris Samaras, and Fusheng Wang.
\newblock A systematic study of key elements underlying molecular property prediction.
\newblock \emph{Nature Communications}, 14\penalty0 (1), 2023.
\newblock URL \url{https://doi.org/10.1038/s41467-023-41948-6}.

\bibitem[Xia et~al.(2023)Xia, Zhang, Zhu, Liu, Gao, Hu, Tan, Zheng, Li, and Li]{Xia2023understandinglimitationsofdeep}
Jun Xia, Lecheng Zhang, Xiao Zhu, Yue Liu, Zhangyang Gao, Bozhen Hu, Cheng Tan, Jiangbin Zheng, Siyuan Li, and Stan~Z. Li.
\newblock Understanding the limitations of deep models for molecular property prediction: Insights and solutions.
\newblock In A.~Oh, T.~Naumann, A.~Globerson, K.~Saenko, M.~Hardt, and S.~Levine, editors, \emph{Advances in Neural Information Processing Systems}, volume~36, pages 64774--64792. Curran Associates, Inc., 2023.
\newblock URL \url{https://proceedings.neurips.cc/paper_files/paper/2023/file/cc83e97320000f4e08cb9e293b12cf7e-Paper-Conference.pdf}.

\bibitem[Cheng et~al.(2024)Cheng, Xiang, Ma, Zeng, Jin, Yang, Lin, Deng, Song, Feng, Deng, and Zeng]{chen2024maskmol}
Zhixiang Cheng, Hongxin Xiang, Pengsen Ma, Li~Zeng, Xin Jin, Xixi Yang, Jianxin Lin, Yang Deng, Bosheng Song, Xinxin Feng, Changhui Deng, and Xiangxiang Zeng.
\newblock {MaskMol}: Knowledge-guided molecular image pre-training framework for activity cliffs with pixel masking.
\newblock \emph{bioRxiv}, 2024.
\newblock URL \url{https://doi.org/10.1101/2024.09.04.611324}.

\bibitem[Iqbal et~al.(2021)Iqbal, Vogt, and Bajorath]{iqbal2021prediction}
Javed Iqbal, Martin Vogt, and J{\"u}rgen Bajorath.
\newblock Prediction of activity cliffs on the basis of images using convolutional neural networks.
\newblock \emph{Journal of Computer-Aided Molecular Design}, pages 1--8, 2021.
\newblock URL \url{https://doi.org/10.1007/s10822-021-00380-y}.

\bibitem[Park et~al.(2022)Park, Sung, Lee, Kang, and Park]{park_acgcn}
Junhui Park, Gaeun Sung, SeungHyun Lee, SeungHo Kang, and ChunKyun Park.
\newblock {ACGCN}: Graph convolutional networks for activity cliff prediction between matched molecular pairs.
\newblock \emph{Journal of Chemical Information and Modeling}, 62\penalty0 (10):\penalty0 2341--2351, 2022.
\newblock URL \url{https://doi.org/10.1021/acs.jcim.2c00327}.

\bibitem[Horvath et~al.(2016)Horvath, Marcou, Varnek, Kayastha, de~la Vega~de Le\'on, and Bajorath]{horvath2016prediction}
Dragos Horvath, Gilles Marcou, Alexandre Varnek, Shilva Kayastha, Antonio de~la Vega~de Le\'on, and J\"urgen Bajorath.
\newblock Prediction of activity cliffs using condensed graphs of reaction representations, descriptor recombination, support vector machine classification, and support vector regression.
\newblock \emph{Journal of Chemical Information and Modeling}, 56\penalty0 (9):\penalty0 1631--1640, 2016.
\newblock URL \url{https://doi.org/10.1021/acs.jcim.6b00359}.

\bibitem[Dablander et~al.(2023)Dablander, Hanser, Lambiotte, et~al.]{Dablander2023}
Michael Dablander, Thomas Hanser, Renaud Lambiotte, et~al.
\newblock Exploring qsar models for activity-cliff prediction.
\newblock \emph{Journal of Cheminformatics}, 15:\penalty0 47, 2023.
\newblock \doi{10.1186/s13321-023-00708-w}.
\newblock URL \url{https://doi.org/10.1186/s13321-023-00708-w}.

\bibitem[van Tilborg et~al.(2022)van Tilborg, Alenicheva, and Grisoni]{Tilborg2022moleculeace}
Derek van Tilborg, Alisa Alenicheva, and Francesca Grisoni.
\newblock Exposing the limitations of molecular machine learning with activity cliffs.
\newblock \emph{Journal of Chemical Information and Modeling}, 62\penalty0 (23):\penalty0 5938--5951, 2022.
\newblock URL \url{https://doi.org/10.1021/acs.jcim.2c01073}.

\bibitem[Kulis(2013)]{Kulis2013}
Brian Kulis.
\newblock {Metric Learning: A Survey}.
\newblock \emph{Foundations and Trends in Machine Learning}, 5\penalty0 (4):\penalty0 287--364, 2013.
\newblock URL \url{http://dx.doi.org/10.1561/2200000019}.

\bibitem[Liu(2011)]{Liu2011}
Tie-Yan Liu.
\newblock \emph{Learning to Rank for Information Retrieval}.
\newblock Springer-Verlag Berlin Heidelberg, Berlin, Heidelberg, 1st edition, 2011.
\newblock URL \url{https://doi.org/10.1007/978-3-642-14267-3}.

\bibitem[Altae-Tran et~al.(2017)Altae-Tran, Ramsundar, Pappu, and Pande]{Altae-Tran2017-OneShot}
Han Altae-Tran, Bharath Ramsundar, Aneesh~S. Pappu, and Vijay Pande.
\newblock Low data drug discovery with one-shot learning.
\newblock \emph{ACS Central Science}, 3\penalty0 (4):\penalty0 283--293, 2017.
\newblock URL \url{https://doi.org/10.1021/acscentsci.6b00367}.

\bibitem[Vella and Ebejer(2023)]{Vella2023-FS}
Daniel Vella and Jean-Paul Ebejer.
\newblock Few-shot learning for low-data drug discovery.
\newblock \emph{Journal of Chemical Information and Modeling}, 63\penalty0 (1):\penalty0 27--42, 2023.
\newblock URL \url{https://doi.org/10.1021/acs.jcim.2c00779}.

\bibitem[Stanley et~al.(2021)Stanley, Bronskill, Maziarz, Misztela, Lanini, Segler, Schneider, and Brockschmidt]{Stanley-FsMol}
Megan Stanley, John Bronskill, Krzysztof Maziarz, Hubert Misztela, Jessica Lanini, Marwin Segler, Nadine Schneider, and Marc Brockschmidt.
\newblock {FS-Mol}: A few-shot learning dataset of molecules.
\newblock In J.~Vanschoren and S.~Yeung, editors, \emph{Proceedings of the Neural Information Processing Systems Track on Datasets and Benchmarks}, volume~1, 2021.
\newblock URL \url{https://datasets-benchmarks-proceedings.neurips.cc/paper_files/paper/2021/file/8d3bba7425e7c98c50f52ca1b52d3735-Paper-round2.pdf}.

\bibitem[Tagasovska et~al.(2024)Tagasovska, Gligorijević, Cho, and Loukas]{tagasovska2024implicitlyguideddesignpropen}
Nataša Tagasovska, Vladimir Gligorijević, Kyunghyun Cho, and Andreas Loukas.
\newblock Implicitly guided design with {PropEn}: Match your data to follow the gradient, 2024.
\newblock URL \url{https://arxiv.org/abs/2405.18075}.

\bibitem[Wetzel et~al.(2022{\natexlab{a}})Wetzel, Ryczko, Melko, and Tamblyn]{Wetzel2022-A}
Sebastian~Johann Wetzel, Kevin Ryczko, Roger~Gordon Melko, and Isaac Tamblyn.
\newblock Twin neural network regression.
\newblock \emph{Applied AI Letters}, 3\penalty0 (4), 2022{\natexlab{a}}.
\newblock URL \url{http://dx.doi.org/10.1002/ail2.78}.

\bibitem[Wetzel et~al.(2022{\natexlab{b}})Wetzel, Melko, and Tamblyn]{Wetzel2022-B}
Sebastian~J Wetzel, Roger~G Melko, and Isaac Tamblyn.
\newblock Twin neural network regression is a semi-supervised regression algorithm.
\newblock \emph{Machine Learning: Science and Technology}, 3\penalty0 (4):\penalty0 045007, 2022{\natexlab{b}}.
\newblock URL \url{http://dx.doi.org/10.1088/2632-2153/ac9885}.

\bibitem[Belaid et~al.(2024)Belaid, Rabus, and Hüllermeier]{Belaid2024}
Mohamed~Karim Belaid, Maximilian Rabus, and Eyke Hüllermeier.
\newblock Pairwise difference learning for classification, 2024.
\newblock URL \url{https://arxiv.org/abs/2406.20031}.

\bibitem[Tynes et~al.(2021)Tynes, Gao, Burrill, Batista, Perez, Yang, and Lubbers]{Tynes2021}
Michael Tynes, Wenhao Gao, Daniel~J. Burrill, Enrique~R. Batista, Danny Perez, Ping Yang, and Nicholas Lubbers.
\newblock Pairwise difference regression: A machine learning meta-algorithm for improved prediction and uncertainty quantification in chemical search.
\newblock \emph{Journal of Chemical Information and Modeling}, 61\penalty0 (8):\penalty0 3846--3857, 2021.
\newblock URL \url{https://doi.org/10.1021/acs.jcim.1c00670}.

\bibitem[Fralish et~al.(2023)Fralish, Chen, Skaluba, and Reker]{Fralish2023}
Zachary Fralish, Ashley Chen, Paul Skaluba, and Daniel Reker.
\newblock {DeepDelta}: predicting {ADMET} improvements of molecular derivatives with deep learning.
\newblock \emph{Journal of Cheminformatics}, 15\penalty0 (1):\penalty0 101, 2023.
\newblock URL \url{https://doi.org/10.1186/s13321-023-00769-x}.

\bibitem[Fralish and Reker(2024)]{Fralish2024activelearningpairs}
Zachary Fralish and Daniel Reker.
\newblock Finding the most potent compounds using active learning on molecular pairs.
\newblock \emph{Beilstein Journal of Organic Chemistry}, 20:\penalty0 2152--2162, 2024.
\newblock URL \url{https://doi.org/10.3762/bjoc.20.185}.

\bibitem[Breiman(2001)]{Breiman2001-nj}
Leo Breiman.
\newblock Random forests.
\newblock \emph{Machine Learning}, 45\penalty0 (1):\penalty0 5--32, 2001.
\newblock URL \url{https://doi.org/10.1023/A:1010933404324}.

\bibitem[Chen and Guestrin(2016)]{chen2016xgboost}
Tianqi Chen and Carlos Guestrin.
\newblock {XGBoost}: A scalable tree boosting system.
\newblock In \emph{Proceedings of the 22nd ACM SIGKDD International Conference on Knowledge Discovery and Data Mining}, KDD '16, pages 785--794, New York, NY, USA, 2016. ACM.
\newblock URL \url{http://doi.acm.org/10.1145/2939672.2939785}.

\bibitem[Rogers and Hahn(2010)]{Rogers2010-vf}
David Rogers and Mathew Hahn.
\newblock Extended-connectivity fingerprints.
\newblock \emph{Journal of Chemical Information and Modeling}, 50\penalty0 (5):\penalty0 742--754, 2010.
\newblock URL \url{https://doi.org/10.1021/ci100050t}.

\bibitem[Landrum(2016)]{Landrum2016RDKit2016_09_4}
Greg Landrum.
\newblock {RDKit}: Open-source cheminformatics software, 2016.
\newblock URL \url{https://github.com/rdkit/rdkit/releases/tag/Release_2016_09_4}.

\bibitem[Hu* et~al.(2020)Hu*, Liu*, Gomes, Zitnik, Liang, Pande, and Leskovec]{hu2020gine}
Weihua Hu*, Bowen Liu*, Joseph Gomes, Marinka Zitnik, Percy Liang, Vijay Pande, and Jure Leskovec.
\newblock Strategies for pre-training graph neural networks.
\newblock In \emph{International Conference on Learning Representations}, 2020.
\newblock URL \url{https://openreview.net/forum?id=HJlWWJSFDH}.

\bibitem[Corso et~al.(2020)Corso, Cavalleri, Beaini, Li\`{o}, and Velickovic]{corso2020principalneighbourhoodaggregationgraph}
Gabriele Corso, Luca Cavalleri, Dominique Beaini, Pietro Li\`{o}, and Petar Velickovic.
\newblock Principal neighbourhood aggregation for graph nets.
\newblock In \emph{Proceedings of the 34th International Conference on Neural Information Processing Systems}, NIPS '20, Red Hook, NY, USA, 2020. Curran Associates Inc.
\newblock URL \url{https://proceedings.neurips.cc/paper/2020/file/99cad265a1768cc2dd013f0e740300ae-Paper.pdf}.

\bibitem[Wang et~al.(2022)Wang, Wang, Cao, and Barati~Farimani]{wang2022molclr}
Yuyang Wang, Jianren Wang, Zhonglin Cao, and Amir Barati~Farimani.
\newblock Molecular contrastive learning of representations via graph neural networks.
\newblock \emph{Nature Machine Intelligence}, 4\penalty0 (3):\penalty0 279--287, 2022.
\newblock URL \url{http://dx.doi.org/10.1038/s42256-022-00447-x}.

\bibitem[Kaufman et~al.(2024)Kaufman, Williams, Underkoffler, Pederson, Mardirossian, Watson, and Parkhill]{kaufman2023coati}
Benjamin Kaufman, Edward~C. Williams, Carl Underkoffler, Ryan Pederson, Narbe Mardirossian, Ian Watson, and John Parkhill.
\newblock {COATI}: Multimodal contrastive pretraining for representing and traversing chemical space.
\newblock \emph{Journal of Chemical Information and Modeling}, 64\penalty0 (4):\penalty0 1145--1157, 2024.
\newblock URL \url{https://doi.org/10.1021/acs.jcim.3c01753}.

\bibitem[Noutahi et~al.(2023)Noutahi, Gabellini, Craig, Lim, and Tossou]{noutahi2023gottasafenewframework}
Emmanuel Noutahi, Cristian Gabellini, Michael Craig, Jonathan S.~C Lim, and Prudencio Tossou.
\newblock Gotta be {SAFE}: A new framework for molecular design, 2023.
\newblock URL \url{https://arxiv.org/abs/2310.10773}.

\bibitem[Zhou et~al.(2023)Zhou, Gao, Ding, Zheng, Xu, Wei, Zhang, and Ke]{zhou2023unimol}
Gengmo Zhou, Zhifeng Gao, Qiankun Ding, Hang Zheng, Hongteng Xu, Zhewei Wei, Linfeng Zhang, and Guolin Ke.
\newblock {Uni-Mol}: A universal {3D} molecular representation learning framework.
\newblock In \emph{The Eleventh International Conference on Learning Representations}, 2023.
\newblock URL \url{https://openreview.net/forum?id=6K2RM6wVqKu}.

\bibitem[Ehrlich and Rarey(2012)]{ehrlich2012}
Hans-Christian Ehrlich and Matthias Rarey.
\newblock Systematic benchmark of substructure search in molecular graphs - from {Ullmann} to {VF2}.
\newblock \emph{Journal of Cheminformatics}, 4\penalty0 (1):\penalty0 13, 2012.
\newblock URL \url{https://doi.org/10.1186/1758-2946-4-13}.

\bibitem[Pedregosa et~al.(2011)Pedregosa, Varoquaux, Gramfort, Michel, Thirion, Grisel, Blondel, Prettenhofer, Weiss, Dubourg, Vanderplas, Passos, Cournapeau, Brucher, Perrot, and Duchesnay]{scikit-learn}
F.~Pedregosa, G.~Varoquaux, A.~Gramfort, V.~Michel, B.~Thirion, O.~Grisel, M.~Blondel, P.~Prettenhofer, R.~Weiss, V.~Dubourg, J.~Vanderplas, A.~Passos, D.~Cournapeau, M.~Brucher, M.~Perrot, and E.~Duchesnay.
\newblock Scikit-learn: Machine learning in {P}ython.
\newblock \emph{Journal of Machine Learning Research}, 12\penalty0 (85):\penalty0 2825--2830, 2011.
\newblock URL \url{http://jmlr.org/papers/v12/pedregosa11a.html}.

\end{thebibliography}
}


\appendix

\section{Appendix}
\label{appendix}

\subsection{Models}
\label{appendix:models}

We used the following models to evaluate the effectiveness of the SQRL approach:

\paragraph{Baselines.} All baseline models were trained on top of Morgan count fingerprints of size 2048, radius 2, and including chirality. For RF, XGBoost, and KNN, RDKit features from~\citep{Landrum2016RDKit2016_09_4} were concatenated to Morgan fingeprints.
\begin{itemize}
    \item \textbf{Random Forest (RF)}: An ensemble learning method using decision trees. Scikit-learn implementation was used~\citep{scikit-learn} with default parameters.
    \item \textbf{XGBoost}: A gradient boosting framework, optimized for efficiency and performance. Implementation from Ref.~\citenum{chen2016xgboost} was used with default parameters.
    \item \textbf{k-Nearest Neighbors (KNN)}: A non-parametric method based on the similarity between data points. Scikit-learn implementation was used~\citep{scikit-learn} with $k=1$ to compare with our relative setting where we evaluate with respect to the closest training data point.
    \item \textbf{Multi-Layer Perceptron (MLP)}: A standard feedforward neural network.
\end{itemize}

\paragraph{Graph neural networks.} GNNs were trained end-to-end, with their learned representations followed by MLP layers for standard or relative predictions.
\begin{itemize}
    \item \textbf{AttentiveFP}: A graph neural network that uses graph attention mechanisms to capture atomic interactions~\citep{Xiong2020AttentiveFP}.
    \item \textbf{GINE}: Graph Isomorphism Network with Edge features, enhancing the model's ability to capture bond information~\citep{hu2020gine}.
    \item \textbf{PNA}: Principal Neighbourhood Aggregation, a GNN architecture designed to be stable under permutations~\citep{corso2020principalneighbourhoodaggregationgraph}. The implementation used in this paper follows Ref.~\citenum{Stanley-FsMol}.
    \item \textbf{MolCLR}: A pre-trained GNN using contrastive learning for molecular representation. In our experiments, we did not freeze the MolCLR layers, allowing them to be fine-tuned along with the rest of the model, according to the procedure of the original authors~\citep{wang2022molclr}.
\end{itemize}

\paragraph{Transformer-based models.}
For transfomer-based models, we utilized the pre-trained embeddings as fixed feature extractors, followed by MLP for task-specific predictions.
\begin{itemize}
    \item \textbf{SAFE-GPT}: A chemical language model trained on linear molecular notation which has been adapted for autoregressive tasks~\citep{noutahi2023gottasafenewframework}.
    \item \textbf{Uni-Mol}: A 3D-aware transformer model trained on molecular conformations using SE(3) equivariant operations~\citep{zhou2023unimol}. Due to computationally expensive conformation generation, 
    \item \textbf{COATI}: A multi-modal generative model combining 2D and 3D molecular information through contrastive learning~\citep{kaufman2023coati}. To obtain embeddings for our applications, the text encoder part of the model was used (Barlow\_Closed).
\end{itemize}

Due to the computationally expensive conformer generation step, Uni-Mol-SQRL was evaluated on the following subset of MoleculeACE tasks (23/30): \texttt{CHEMBL2971\_Ki}, \texttt{CHEMBL2835\_Ki}, \texttt{CHEMBL219\_Ki}, \texttt{CHEMBL228\_Ki}, \texttt{CHEMBL238\_Ki}, \texttt{CHEMBL1862\_Ki}, \texttt{CHEMBL218\_EC50}, \texttt{CHEMBL231\_Ki}, \texttt{CHEMBL235\_EC50}, \texttt{CHEMBL287\_Ki}, \texttt{CHEMBL2147\_Ki}, \texttt{CHEMBL2047\_EC50}, \texttt{CHEMBL4203\_Ki}, \texttt{CHEMBL2034\_Ki}, \texttt{CHEMBL1871\_Ki}, \texttt{CHEMBL4792\_Ki}, \texttt{CHEMBL244\_Ki}, \texttt{CHEMBL234\_Ki}, \texttt{CHEMBL239\_EC50}, \texttt{CHEMBL262\_Ki}, \texttt{CHEMBL4616\_EC50}, \texttt{CHEMBL3979\_EC50}, \texttt{CHEMBL4005\_Ki}.

\subsection{Hyperparameter selection}
\label{appendix:hpo}

See Table~\ref{appendix:hpo_table} for optimized hyperparameters used for each model in a standard and SQRL setting.

\begin{table}[H]%
\centering%
\caption{Model hyperparameters.}%
\label{appendix:hpo_table}%
\resizebox{\textwidth}{!}{%
\begin{tabular}{lcccccc}%
\toprule
Model & Linear Sizes & GNN Layers & Dropout & Learning Rate & Batch Size & Other \\
\midrule
MLP & [256, 256] & - & 0.0 & 1e-4 & 128 & - \\
MLP-SQRL & [512, 256] & - & 0.2 & 1e-5 & 64 & - \\
AttentiveFP & [256] & 3 & 0.2 & 1e-4 & 128 & timesteps: 2 \\
AttentiveFP-SQRL & [128] & 3 & 0.0 & 1e-3 & 256 & timesteps: 4 \\
GINE & [128] & 4 & 0.0 & 1e-4 & 64 & - \\
GINE-SQRL & [256, 128] & 5 & 0.0 & 1e-3 & 64 & - \\
PNA & [128] & 4 & 0.0 & 1e-4 & 64 & - \\
PNA-SQRL & [128] & 8 & 0.0 & 1e-5 & 256 & - \\
MolCLR & [512] & - & 0.0 & 1e-4 & 128 & - \\
MolCLR-SQRL & [128] & - & 0.0 & 1e-4 & 128 & - \\
COATI & [256] & - & 0.0 & 1e-5 & 64 & - \\
COATI-SQRL & [256, 128] & - & 0.0 & 1e-3 & 128 & - \\
SAFE-GPT & [256, 128] & - & 0.0 & 1e-3 & 32 & - \\
SAFE-GPT-SQRL & [256, 128] & - & 0.0 & 1e-4 & 128 & - \\
Uni-Mol & [256, 128] & - & 0.0 & 1e-3 & 128 & - \\
Uni-Mol-SQRL & [128] & - & 0.0 & 1e-4 & 128 & - \\
\bottomrule
\end{tabular}%
}%
\end{table}%

\newpage
\subsection{MoleculeACE Dataset}

\begin{figure}[H]%
    \centering%
    \includegraphics[width=\linewidth]{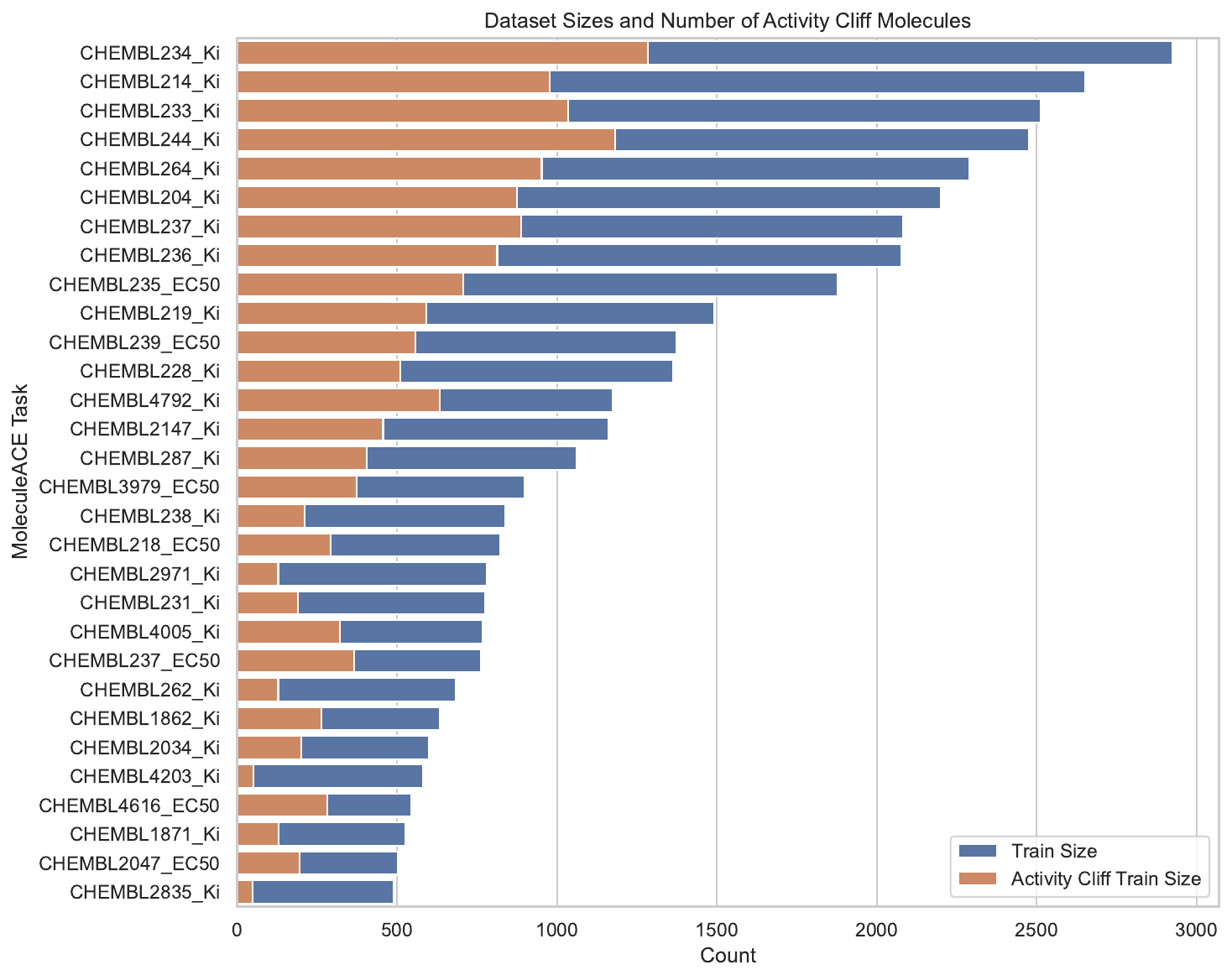}%
    \caption{Training data sizes for each task in MoleculeACE.}%
    \label{fig:moleculeace_tasks_data_sizes}%
\end{figure}

The MoleculeACE dataset provided by \citeauthor{Tilborg2022moleculeace} is curated from ChEMBL v29 and contains potency measurements for 30 targets. Activity cliff molecules were defined as pairs of molecules with greater than 90\% substructure, scaffold, or SMILES similarity and greater than 10-fold activity difference. Figure~\ref{fig:moleculeace_tasks_data_sizes} shows the number of training samples for each task in the MoleculeACE dataset along with the number of activity cliff molecules present in each task. The train-test split was performed by clustering molecules into 5 clusters and stratified splitting using the activity cliff label.

\label{appendix:moleculeace_tasks}

\newpage
\subsection{Examples of molecular pairs}
\label{appendix:pairs}

\begin{figure}[H]%
    \centering%
    \begin{minipage}[t]{0.1\linewidth}%
        \flushleft
        \begin{TAB}(r)[0pt,0.1\linewidth,0.85\textheight]{l}{ccccccc}%
            $\alpha=0.1$ \\ $\alpha=0.2$ \\ $\alpha=0.3$ \\ $\alpha=0.4$ \\ $\alpha=0.5$ \\ $\alpha=0.6$ \\ $\alpha=0.7$ \\
        \end{TAB}%
    \end{minipage}\hfill%
    \begin{minipage}[t]{0.41\linewidth}%
        \centering%
        \includegraphics[height=0.85\textheight]{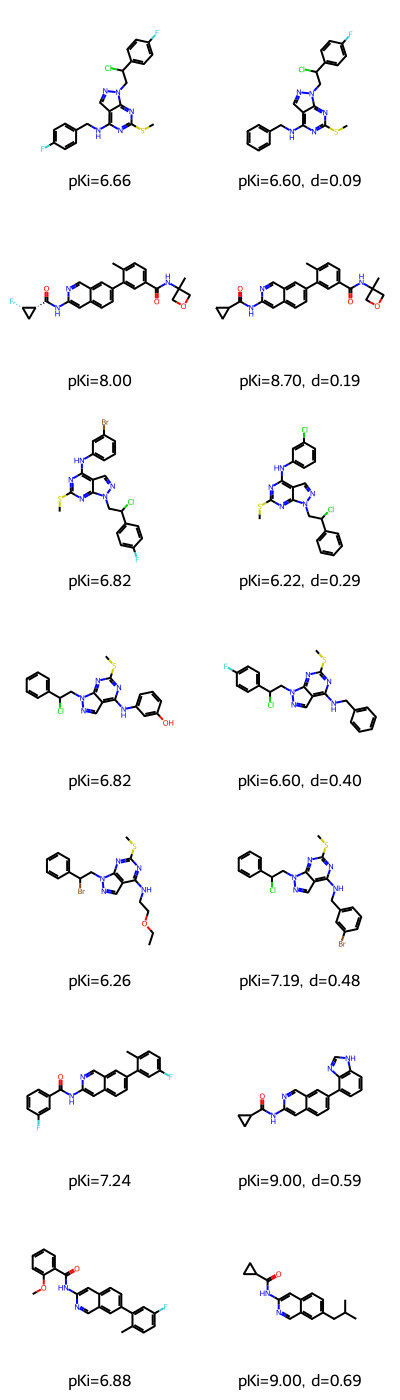}%
        \caption{Molecular pairs obtained by the data matching procedure described in Section~\ref{sec:methods} at different Tanimoto distance thresholds $\alpha$ for MoleculeACE task \texttt{CHEMBL1862\_Ki}.}%
        \label{fig:normal_pairs}%
    \end{minipage}\hfill%
    \begin{minipage}[t]{0.41\linewidth}%
        \centering%
        \includegraphics[height=0.85\textheight]{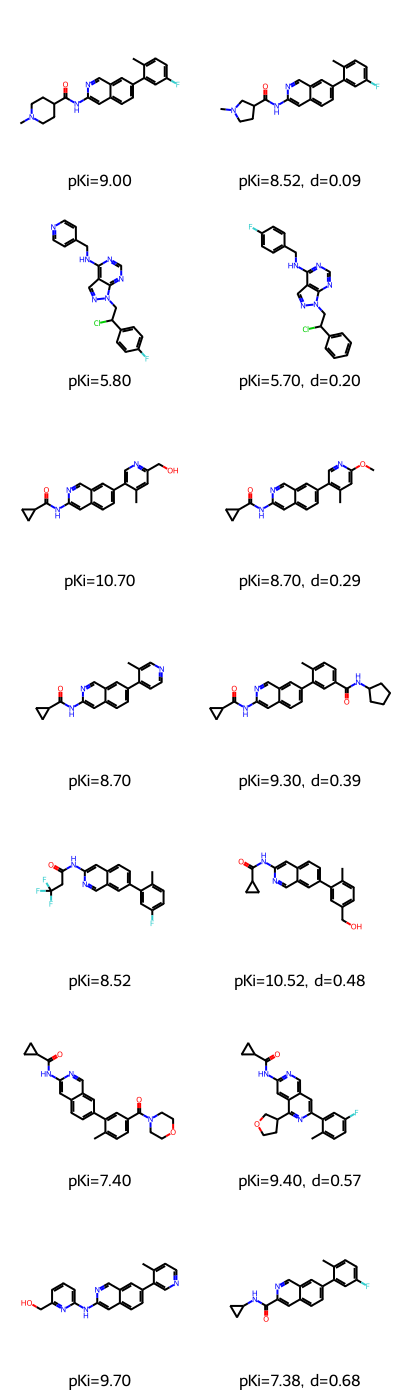}%
        \caption{Molecular pairs of \emph{activity cliff molecules} obtained by the data matching procedure described in Section~\ref{sec:methods} at different Tanimoto distance thresholds $\alpha$ for MoleculeACE task \texttt{CHEMBL1862\_Ki}.}%
        \label{fig:cliff_pairs}%
    \end{minipage}%
\end{figure}%

\newpage
\subsection{Additional results}
\label{appendix:dist_sweep}

\begin{figure}[H]%
    \centering%
    \includegraphics[width=\linewidth]{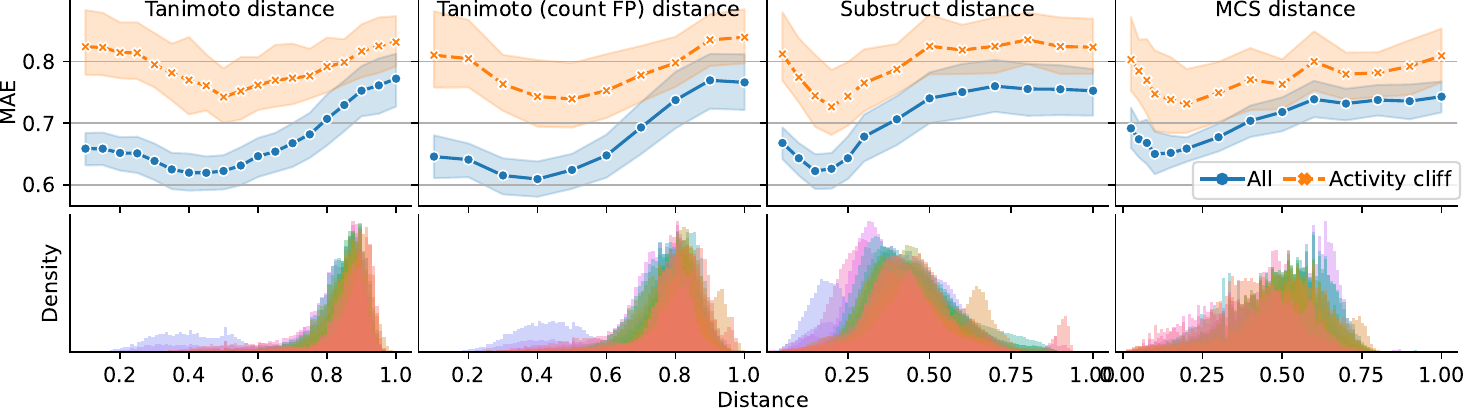}\vspace{6pt}
    \includegraphics[width=0.75\linewidth]{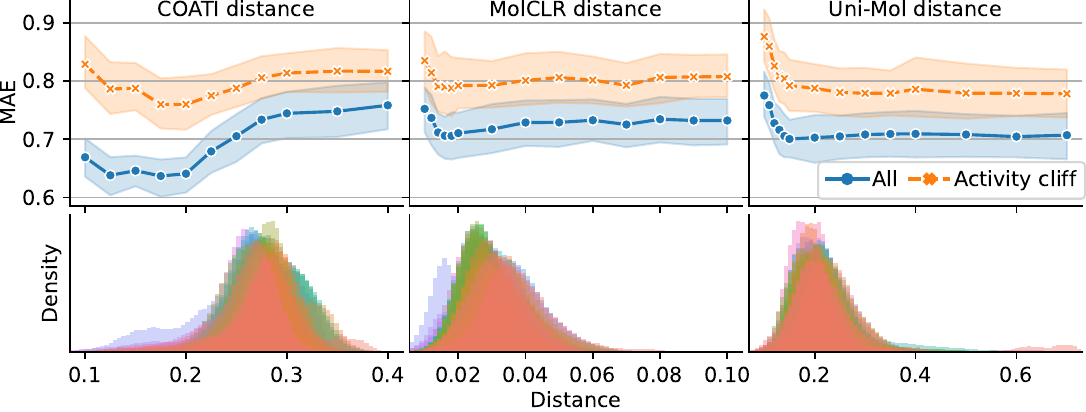}%
    \caption{\textbf{Leveraging local structural information enhances predictive performance.} MAE ($\downarrow$) as a function of distance threshold $\alpha$ for several additional distance metrics compared to Figure~\ref{fig:dist_sweep}, as well as pairwise distance distributions for each metric.
    \emph{Tanimoto}: Tanimoto (Jaccard) distance between binary Morgan fingerprints.
    \emph{Tanimoto (count FP)}: Tanimoto (Jaccard) distance between count-based Morgan fingerprints.
    \emph{Substruct}: Tanimoto (Jaccard) distance between substructure count vectors using a list of 1242 predefined substructures from \citet{ehrlich2012}. \emph{MCS}: Distance metric based on maximum common substructure (MCS) defined as $1 - 2 N_\text{MCS} / (N_i + N_j)$ where $N_\text{MCS}$ is the number of atoms in the MCS, $N_i$ is the number of atoms in molecule $i$, and $N_j$ is the number of atoms in molecule $j$. \emph{COATI~\citep{kaufman2023coati}, MolCLR~\citep{wang2022molclr}, Uni-Mol~\citep{zhou2023unimol}}: Euclidean distances between neural network embeddings obtained with these pre-trained models.}%
    \label{fig:dist_sweep_appendix}%
\end{figure}%

\clearpage

\begin{table}[H]
\centering
\caption{\textbf{Mean Absolute Error (MAE) ($\downarrow$) comparison of predictive performance.} MAE with standard deviation across tasks for Standard and SQRL methods using Tanimoto distance with $\alpha = 0.7$. \textbf{Bold values} indicate the better performing method (lower MAE) for each model and dataset. Uni-Mol was evaluated on a smaller subset of tasks due to computational constraints (see Appendix~\ref{appendix:models}.)}%
\label{tab:mae_results}
\setlength{\tabcolsep}{2pt}
\begin{tabularx}{\linewidth}{@{}lYYYYYY@{}}
\toprule
& \multicolumn{2}{c}{\textbf{MoleculeACE}} & \multicolumn{2}{c}{\textbf{MoleculeACE-Cliff}} & \multicolumn{2}{c}{\textbf{Internal Targets}} \\
\cmidrule(lr){2-3} \cmidrule(lr){4-5} \cmidrule(lr){6-7}
\textbf{Model} & Standard & SQRL & Standard & SQRL & Standard & SQRL \\
\midrule
\emph{Baselines} \\
\cmidrule(r){1-1}
XGBoost & \textbf{0.54 $\pm$ 0.07} & 0.59 $\pm$ 0.08 & \textbf{0.64 $\pm$ 0.12} & 0.71 $\pm$ 0.13 & \textbf{0.41 $\pm$ 0.18} & 0.55 $\pm$ 0.19 \\
RF & \textbf{0.54 $\pm$ 0.07} & 0.57 $\pm$ 0.08 & \textbf{0.63 $\pm$ 0.10} & 0.68 $\pm$ 0.12 & \textbf{0.41 $\pm$ 0.18} & 0.50 $\pm$ 0.21 \\
KNN & \textbf{0.69 $\pm$ 0.12} & 0.77 $\pm$ 0.13 & \textbf{0.83 $\pm$ 0.15} & 0.88 $\pm$ 0.18 & \textbf{0.71 $\pm$ 0.37} & 0.95 $\pm$ 0.35 \\
MLP & 0.94 $\pm$ 0.22 & \textbf{0.67 $\pm$ 0.10} & 0.96 $\pm$ 0.20 & \textbf{0.79 $\pm$ 0.12} & 0.67 $\pm$ 0.41 & \textbf{0.65 $\pm$ 0.20} \\
\midrule
\emph{GNNs} \\
\cmidrule(r){1-1}
AttentiveFP & 0.80 $\pm$ 0.12 & \textbf{0.58 $\pm$ 0.11} & 0.86 $\pm$ 0.12 & \textbf{0.65 $\pm$ 0.16} & 0.56 $\pm$ 0.24 & \textbf{0.50 $\pm$ 0.15} \\
GINE & 1.14 $\pm$ 0.24 & \textbf{0.64 $\pm$ 0.09} & 1.23 $\pm$ 0.32 & \textbf{0.72 $\pm$ 0.14} & 0.75 $\pm$ 0.13 & \textbf{0.47 $\pm$ 0.19} \\
PNA & 0.82 $\pm$ 0.21 & \textbf{0.65 $\pm$ 0.09} & 0.85 $\pm$ 0.18 & \textbf{0.75 $\pm$ 0.14} & 0.63 $\pm$ 0.27 & \textbf{0.58 $\pm$ 0.26} \\
MolCLR & 0.91 $\pm$ 0.12 & \textbf{0.62 $\pm$ 0.09} & 0.95 $\pm$ 0.13 & \textbf{0.72 $\pm$ 0.14} & 0.62 $\pm$ 0.39 & \textbf{0.50 $\pm$ 0.21} \\
\midrule
\emph{Transformers} \\
\cmidrule(r){1-1}
COATI & 0.71 $\pm$ 0.10 & \textbf{0.65 $\pm$ 0.09} & \textbf{0.74 $\pm$ 0.20} & 0.75 $\pm$ 0.14 & \textbf{0.59 $\pm$ 0.30} & 0.61 $\pm$ 0.20 \\
Uni-Mol & 0.94 $\pm$ 0.26 & \textbf{0.69 $\pm$ 0.09} & 0.92 $\pm$ 0.22 & \textbf{0.77 $\pm$ 0.14} & 0.62 $\pm$ 0.41 & \textbf{0.49 $\pm$ 0.48} \\
SAFE-GPT & 0.76 $\pm$ 0.12 & \textbf{0.68 $\pm$ 0.11} & \textbf{0.78 $\pm$ 0.10} & 0.80 $\pm$ 0.13 & \textbf{0.55 $\pm$ 0.22} & 0.60 $\pm$ 0.26 \\
\bottomrule
\end{tabularx}
\end{table}

\begin{figure}[H]%
    \centering%
    \includegraphics[width=1\linewidth]{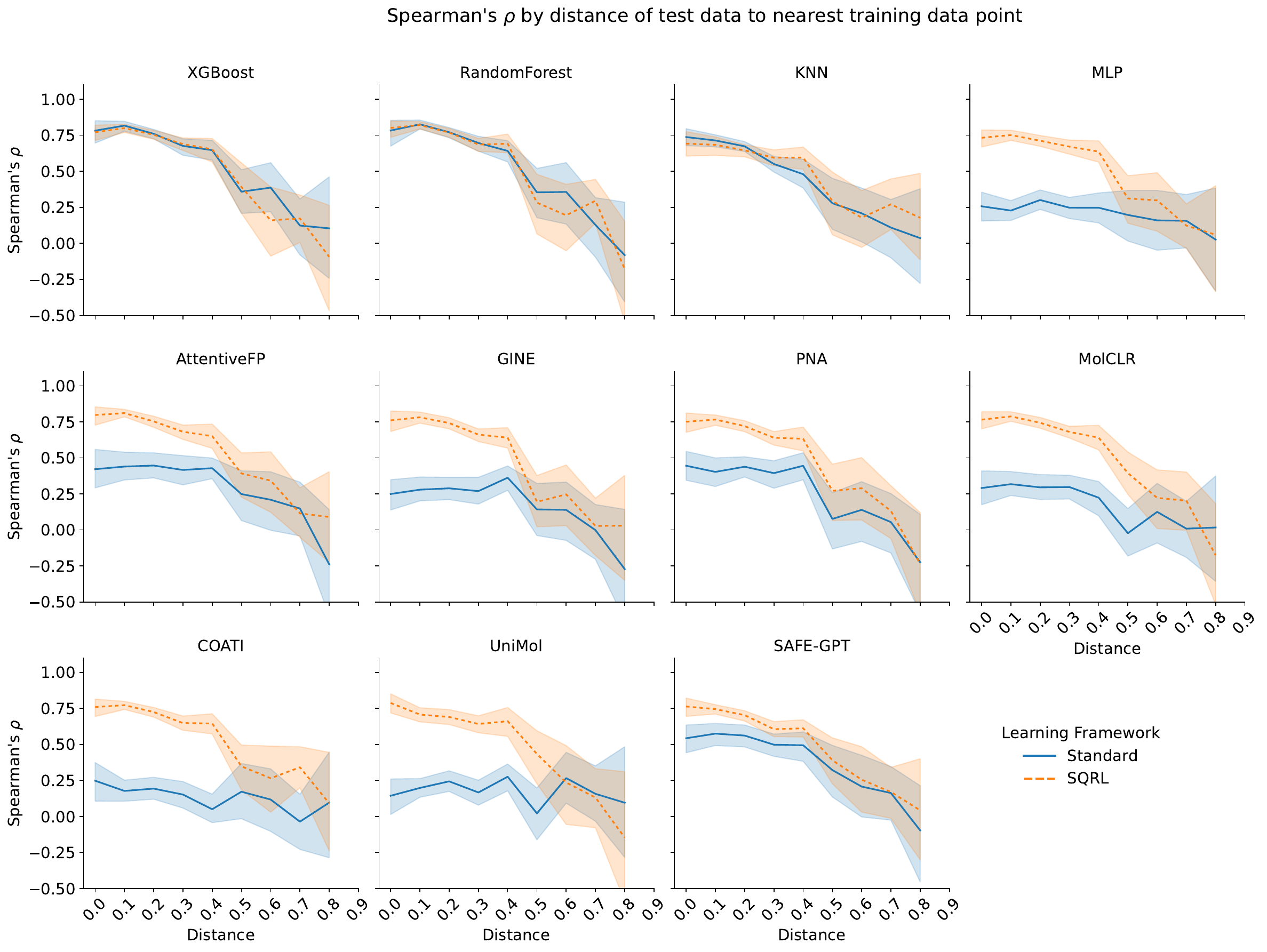}%
    \caption{\textbf{SQRL enhances local molecular consistency.} Spearman's rank correlation coefficient ($\uparrow$) plotted as a function of the distance between test points and their nearest neighbors in the training set. SQRL-trained models that benefit from this training strategy demonstrate the most significant performance gains for test points with close neighbors, while generally maintaining comparable performance to standard-trained models for more distant points. Models in this plot were trained with the distance threshold $\alpha=0.7$ using Tanimoto distance.}%
    \label{fig:dist_to_train}%
\end{figure}%

\end{document}